\documentclass[]{spie}  

\usepackage{amsmath,amsfonts,amssymb}
\usepackage{graphicx}
\usepackage[colorlinks=true, allcolors=blue]{hyperref}

\title{Novel adaptation of video segmentation to 3D MRI: efficient zero-shot knee segmentation with SAM2}

\author[1,2]{Andrew Seohwan Yu}
\author[1]{Mohsen Hariri}
\author[1]{Xuecen Zhang}
\author[2]{Mingrui Yang}
\author[1]{Vipin Chaudhary}
\author[1,2]{Xiaojuan Li}
\affil[1]{Case Western Reserve University, Cleveland, OH, USA}
\affil[2]{Cleveland Clinic, Cleveland, OH, USA}

\authorinfo{Further author information: Andrew Seohwan Yu: asy51@case.edu}

\begin{document} 
\maketitle

\section{Description of purpose}
\label{sec:intro}

Medical image segmentation plays a crucial role in clinical practice by enabling precise diagnosis, treatment planning, and disease monitoring. Despite its importance, traditional segmentation models typically require extensive labeling and training on specific datasets and only target specific imaging modalities or disease types, limiting their generalizability across diverse medical imaging tasks. Segment Anything Model\cite{sam} (SAM1) addresses this limitation by introducing a promptable image segmentation approach. With interactive user inputs such as points and bounding boxes, SAM1 enables precise and context-sensitive segmentation, thus enhancing its versatility. Trained on the large-scale SA-1B dataset, which supports flexible cues for zero-shot segmentation, SAM1 shows excellent generalization capabilities, making it robust across different image types and applications. Recent studies have also further demonstrated the scalability and effectiveness of SAM1 in medical image segmentation.\cite{cheng2023sammed2d,MedSAM,mazurowski2023segment,wu2023medical}

However, with the rapid growth of multimedia content, a large portion of visual data is now recorded in the temporal dimension, especially in video data. A generalized visual segmentation system should be applicable to both images and videos. Based on this, Segment Anything Model 2(SAM2) by Ravi et al\cite{sam2} extends the segmentation capabilities of SAM1 to apply not only to images but also to videos. SAM2 does this by introducing the Promptable Visual Segmentation (PVS) task, which allows the user to define regions of interest by pointing, boxing, or masking on any video frame, and can iteratively update them in subsequent frames. SAM2's streaming memory mechanism allows the model to store and utilize information from previous frames on a frame-by-frame basis as it processes the video, maintaining context and improving segmentation accuracy. Compared to SAM1, SAM2 utilizes a more efficient transformer architecture Hiera\cite{ryali2023hiera}, which allows the model to be up to 6x faster and more accurate in image segmentation tasks. The final large-scale video segmentation dataset (SA-V) includes 35.5M masks covering 50.9K videos, providing rich training and evaluation resources.

\begin{figure}[ht]
\begin{center}
\includegraphics[width=\linewidth]{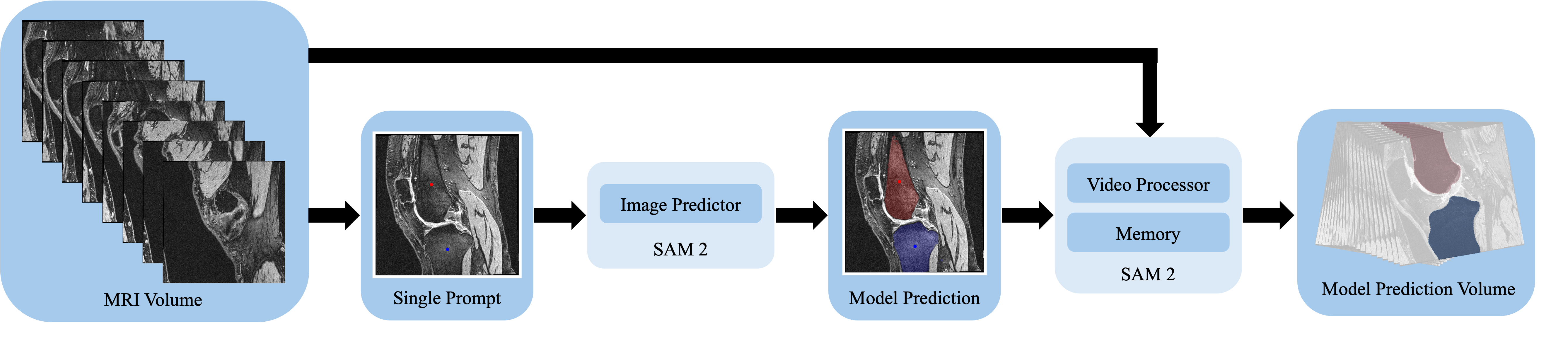}
\end{center}
\caption{\label{fig:sam2} Adaptation of the SAM2 model's video predictor to a 3D MR volume in two stages: one slice from the input MRI volume is segmented using the image predictor; then the video processor and memory component propagates the prompts and the predictions to the rest of the volume iteratively.}
\end{figure}

In our work, we extend SAM2 to 3D MRI images by treating slices from 3D volumes as individual video frames. As shown in Figure \ref{fig:sam2}, this method leverages SAM2's video segmentation capabilities to process 3D data efficiently, enabling precise segmentation of volumetric medical images with minimal user input. This extension significantly enhances SAM2's performance in medical imaging tasks, achieving higher accuracy and efficiency. Specifically, for zero-shot single-prompt segmentation of knee MRI images, SAM2 achieves accurate results with minimal interaction, thus improving the overall workflow of medical image analysis.

\section{METHODS}
\label{sec:methods}


Knee MR volumes from the Osteoarthritis Initiative (OAI) were collected, alongside segmentations of the femur and tibia provided by Zuse Institute Berlin (ZIB). The 488 volumes, each with a matrix size of \(160 \times 384 \times 384\), were acquired using the 3D double-echo steady state (DESS) sequence, automatically segmented combining statistical shape modeling, then manually corrected to produce the OAI-ZIB dataset\cite{ambellan2019automated}. The corrected bone masks were used as ground truth targets in this study.

This study focused on the usage of point prompts to direct SAM1 and SAM2's predictions. Using the ground truth masks of the femur and tibia, the center of each bone's mass was computed for each slice of the volume. The density of the bone was assumed to be uniform throughout the slice, and thus the computed center of mass was equivalent to the centroid. In addition to the centroids, one more point prompt per volume was manually generated, pointing to the patella on the middle (80th) slice. With two point prompts per slice and 160 slices per volume, plus one prompt for the patella, a total of 321 point prompts were generated for each volume. This study also used bounding boxes, generated using the same ground truth bone masks. The min and max \(x\) and \(y\) values of each bone mask in each slice were used to delineate the square region containing the bone. With four points per box, two boxes per slice, and 160 slices per volume, a total of 1280 points were generated for each volume.


SAM1's primary innovation lies in its promptable segmentation tasks, wherein the model generates valid segmentation masks based on various prompts, such as points, bounding boxes, or zero-shot methods, effectively handling ambiguities and generalizing across diverse downstream tasks. The model is trained on the SA-1B dataset, the largest of its kind, containing over 1 billion masks across 11 million images.


SAM1's architecture comprises three main components: an image encoder, a flexible prompt encoder, and a mask decoder. The image encoder, based on the Vision Transformer (ViT), processes input images into lower-dimensional embedding space. The prompt encoder accommodates both sparse (point) and dense (bounding box) prompts, while the mask decoder generates the final segmentation masks from these embeddings. Although SAM1 has been trained on general datasets rather than domain-specific datasets (e.g., medical images), prompting techniques enable its application to domain-specific tasks, such as the annotation of organs and structures in the human body.

Three prompt schemes can be used with SAM1: a) automatic mask generation, b) points prompting, and c) bounding box prompting. With automatic mask generation, no prompts are provided and SAM1 segments the entire image into masked components, some of which may correspond to the objects of interest. This scheme did not align with the needs of the project and was omitted. With point and bounding box prompting, the aforementioned 320 centroids and 1280 points, respectively, were used to generate 320 mask predictions, for each bone in each slice of the volume.

In all, six configurations of SAM1 were tested, including the point and box prompt schemes and three model sizes: base, large, and huge.





SAM2 claimed its superiority to SAM1 based on two major factors: (1) a more efficient and powerful network called Hiera; and (2) new components in the model that enables memory across frames of a video. We tested the first claim by comparing the performance of SAM1 to SAM2's memory-less image segmentation, using the same per-slice point prompt scheme. 

The second claim was tested by activating SAM2's memory-based video segmentation and using two different prompt schemes: (1) a single positive point prompt per bone at the middle (80th) slice of the volume; and (2) single positive point prompt per bone plus two negative point prompts for the other bones (two of femur, tibia, and patella) at the middle slice, resulting in 3 point prompts per bone per volume.

In all, twelve configurations of SAM2 were tested, including the point, point + video, and 3 points + video prompt schemes, and four model sizes: tiny, small, large, base-plus, and large.

\begin{figure}[ht]
\begin{center}
\includegraphics[width=0.32\textwidth]{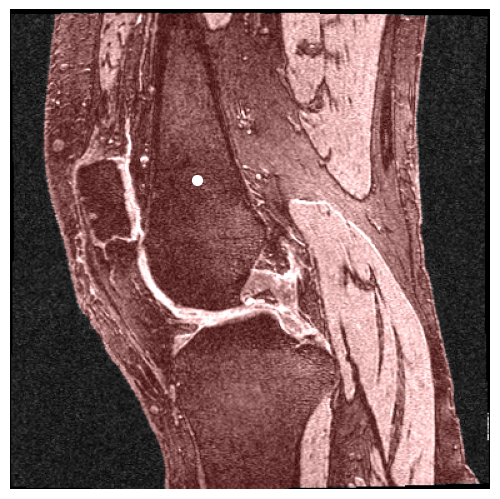}
\includegraphics[width=0.32\textwidth]{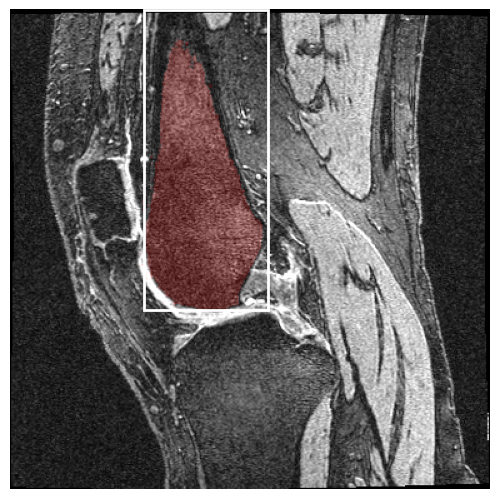}
\includegraphics[width=0.32\textwidth]{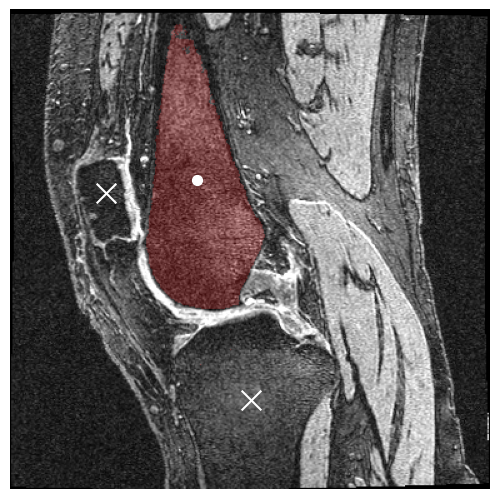}
\end{center}
\caption{\label{fig:prompts} Different prompts (white) provided to SAM1 and SAM2, with femur segmentation predictions overlaid (red). Single point prompt fails to limit the extent of the prediction to the femur (left); bounding box (center) and negative prompts (right) successfully limit the extent.}
\end{figure}

\section{RESULTS}
\label{sec:results}

The performance of the SAM1 and SAM2 models were evaluated by comparing the model predictions to the ground truth bone segmentation masks. Dice similarity coefficient (DSC) \cite{sorensen1948method}, a common image segmentation metric, was used in this study. For each model and prompt scheme, three median DSC scores were reported: femur only, tibia only, and femur and tibia combined. Finally, the number of parameters for each variant of both models were noted as well, as shown in Tables \ref{tab:sam1dice}. 

For SAM1, the results showed a general trend in which bounding box prompts outperformed point prompts and larger models performed better. Neither of these results were surprising, especially since bounding boxes provide more information to the model than points. In particular, bounding boxes dictate the size of the object of interest, unlike points. As shown in Figure \ref{fig:prompts}, a common failure case with point prompts was the model's tendency to select the entire foreground. On the other hand, the positive correlation between model size and DSC was minimal.

For SAM2, the results were more unexpected. Regarding model sizes, the base-plus model was on par with or outperformed the large model, despite having nearly four times fewer parameters. The tiny and small model also had surprisingly high DSC on certain tasks. For the most direct comparison between SAM1 and SAM2, wherein both models used a point prompt for each slice of the MR volume, the DSC scores were about the same. That is, the SAM2 underperformed relative to expectations. When SAM2's video predictor was activated with the ``\(\text{3 points}+\text{video}\)" prompt scheme, the model boasted its highest and median DSC at 0.9196. Note that the 
``point" prompt scheme provided two points (femur and tibia) for each slice of the 160-slice volume, resulting in 320 total prompts per volume. Providing only three points proved to be the better option for SAM2. Even the ``\(\text{point}+\text{video}\)" prompt scheme with a single prompt per volume gave results on par with that of the ``point" prompt scheme.

\begin{table}[ht]
\caption{Median testing 3D DSC for SAM1 and SAM2 pretrained models. Point and bounding box prompts are traditional, per-image schemes, whereas point + video prompts leverage SAM2's innovations. More details can be found in Section \ref{sec:methods}.}
\label{tab:sam1dice}
\begin{center}       
\begin{tabular}{|r|p{3.5cm}|r|r|r|r|}
\hline
\rule[-1ex]{0pt}{3.5ex} \textbf{Prompt} & \textbf{SAM1 Model} & \textbf{Parameters} & \textbf{Femur} & \textbf{Tibia} & \textbf{Femur + Tibia} \\
\hline
\rule[-1ex]{0pt}{3.5ex} point & sam\_vit\_base & 93.73M & 0.7457 & \textbf{0.7542} & 0.6046 \\
\hline
\rule[-1ex]{0pt}{3.5ex} point & sam\_vit\_large & 312.34M & 0.6538 & 0.7304 & \textbf{0.7282} \\
\hline
\rule[-1ex]{0pt}{3.5ex} point & sam\_vit\_huge & 641.09M & \textbf{0.7508} & 0.7185 & 0.6527 \\
\hline
\hline
\rule[-1ex]{0pt}{3.5ex} bounding box & sam\_vit\_base & 93.73M & 0.7173 & 0.8033 &  \\
\hline
\rule[-1ex]{0pt}{3.5ex} bounding box  & sam\_vit\_large & 312.34M & 0.8197 &  0.8110 &  \\
\hline
\rule[-1ex]{0pt}{3.5ex} bounding box  & sam\_vit\_huge & 641.09M & \textbf{0.8223} & \textbf{0.8104} &  \\
\hline
\hline
\rule[-1ex]{0pt}{3.5ex} \textbf{Prompt} & \textbf{SAM2 Model} & \textbf{Parameters} & \textbf{Femur} & \textbf{Tibia} & \textbf{Femur + Tibia} \\
\hline
\rule[-1ex]{0pt}{3.5ex} point & sam2\_hiera\_tiny & 38.95M & 0.6021 & 0.7395 & 0.6493 \\
\hline
\rule[-1ex]{0pt}{3.5ex} point & sam2\_hiera\_small & 46.04M & 0.6637 & 0.6374 & 0.6496 \\
\hline
\rule[-1ex]{0pt}{3.5ex} point & sam2\_hiera\_base\_plus & 80.83M & \textbf{0.7409} & \textbf{0.8155} & \textbf{0.7664} \\
\hline
\rule[-1ex]{0pt}{3.5ex} point & sam2\_hiera\_large & 224.43M & 0.6961 & 0.5895 & 0.6511 \\
\hline
\hline
\rule[-1ex]{0pt}{3.5ex} point + video & sam2\_hiera\_tiny & 38.95M & 0.2433 & 0.9137 & 0.3275 \\
\hline
\rule[-1ex]{0pt}{3.5ex} point + video & sam2\_hiera\_small & 46.04M & 0.2971 & 0.9187 & 0.3823 \\
\hline
\rule[-1ex]{0pt}{3.5ex} point + video & sam2\_hiera\_base\_plus & 80.83M & 0.7019 & \textbf{0.9643} & \textbf{0.7707} \\
\hline
\rule[-1ex]{0pt}{3.5ex} point + video & sam2\_hiera\_large & 224.43M & \textbf{0.7154} & 0.9052 & 0.5818 \\
\hline
\hline
\rule[-1ex]{0pt}{3.5ex} 3 points + video & sam2\_hiera\_tiny & 38.95M & 0.9085 & 0.9226 & 0.9044 \\
\hline
\rule[-1ex]{0pt}{3.5ex} 3 points + video & sam2\_hiera\_small & 46.04M & 0.8996 & 0.9234 & 0.8978 \\
\hline
\rule[-1ex]{0pt}{3.5ex} 3 points + video & sam2\_hiera\_base\_plus & 80.83M & 0.8633 & \textbf{0.9545} & 0.8173 \\
\hline
\rule[-1ex]{0pt}{3.5ex} 3 points + video & sam2\_hiera\_large & 224.43M & \textbf{0.9322} & 0.9222 & \textbf{0.9196} \\
\hline
\end{tabular}
\end{center}
\end{table}

\section{New or breakthrough work to be presented}

Our work shows that the adaptation of SAM2 to 3D medical image segmentation tasks is a viable option for medical image segmentation, even with minimal prompting and zero-shot inferencing. Furthermore, the results of this study suggest that the use of SAM2's memory component and video predictor are not only feasible but synergistic with 3D medical volumes, since their prompt propagation across slices are superior to that of prompts that were manually provided using ground truth masks. The memory retained across slices (frames) likely conveys more complex information than is possible with point and bounding box prompts.

\section{CONCLUSIONS}
\label{sec:conclusions}

In this study, we used Segment Anything Model 2 (SAM2) for zero-shot, single-prompt segmentation of knee MRI. By leveraging SAM2's advanced capabilities, we eliminated the need for large labeled datasets and extended its video segmentation approach to three-dimensional medical images. Our experiments on the OAI-ZIB dataset demonstrated that SAM2 can efficiently and accurately segment knee MRI scans with minimal user interaction. This breakthrough offers a scalable and cost-effective solution for automated medical image segmentation, streamlining clinical workflows and enhancing precision in diagnosis and treatment planning. The robustness and versatility of SAM2, combined with its efficiency, position it as a transformative tool in medical image analysis. Future work will explore further optimization for various medical imaging modalities and broader clinical applications, aiming to continue advancing the capabilities and impact of SAM2 in healthcare.

\textbf{Note}: This work is NOT being, nor has been, submitted for publication or presentation elsewhere.

\bibliography{report} 

\begin{thebibliography}{1}

\bibitem{sam}
Kirillov, A., Mintun, E., Ravi, N., Mao, H., Rolland, C., Gustafson, L., Xiao, T., Whitehead, S., Berg, A.~C., Lo, W.-Y., Doll{\'a}r, P., and Girshick, R., ``Segment anything,'' {\em arXiv:2304.02643}  (2023).

\bibitem{cheng2023sammed2d}
Cheng, J., Ye, J., Deng, Z., Chen, J., Li, T., Wang, H., Su, Y., Huang, Z., Chen, J., Sun, L. J.~H., He, J., Zhang, S., Zhu, M., and Qiao, Y., ``Sam-med2d,'' (2023).

\bibitem{MedSAM}
Ma, J., He, Y., Li, F., Han, L., You, C., and Wang, B., ``Segment anything in medical images,'' {\em Nature Communications}~{\bf 15},  1--9 (2024).

\bibitem{mazurowski2023segment}
Mazurowski, M.~A., Dong, H., Gu, H., Yang, J., Konz, N., and Zhang, Y., ``Segment anything model for medical image analysis: an experimental study,'' {\em Medical Image Analysis}~{\bf 89},  102918 (2023).

\bibitem{wu2023medical}
Wu, J., Ji, W., Liu, Y., Fu, H., Xu, M., Xu, Y., and Jin, Y., ``Medical sam adapter: Adapting segment anything model for medical image segmentation,'' (2023).

\bibitem{sam2}
Ravi, N., Gabeur, V., Hu, Y.-T., Hu, R., Ryali, C., Ma, T., Khedr, H., R{\"a}dle, R., Rolland, C., Gustafson, L., Mintun, E., Pan, J., Alwala, K.~V., Carion, N., Wu, C.-Y., Girshick, R., Doll{\'a}r, P., and Feichtenhofer, C., ``Sam 2: Segment anything in images and videos,'' {\em arXiv preprint arXiv:2408.00714}  (2024).

\bibitem{ryali2023hiera}
Ryali, C., Hu, Y.-T., Bolya, D., Wei, C., Fan, H., Huang, P.-Y., Aggarwal, V., Chowdhury, A., Poursaeed, O., Hoffman, J., Malik, J., Li, Y., and Feichtenhofer, C., ``Hiera: A hierarchical vision transformer without the bells-and-whistles,'' {\em ICML}  (2023).

\bibitem{ambellan2019automated}
Ambellan, F., Tack, A., Ehlke, M., and Zachow, S., ``Automated segmentation of knee bone and cartilage combining statistical shape knowledge and convolutional neural networks: Data from the osteoarthritis initiative,'' {\em Medical image analysis}~{\bf 52},  109--118 (2019).

\bibitem{sorensen1948method}
Sorensen, T., ``A method of establishing groups of equal amplitude in plant sociology based on similarity of species content and its application to analyses of the vegetation on danish commons,'' {\em Biologiske skrifter}~{\bf 5},  1--34 (1948).

\end{thebibliography}
\bibliographystyle{spiebib} 

\end{document}